\documentclass{article}

\usepackage{microtype}
\usepackage{graphicx}
\usepackage{chngcntr}
\usepackage{subcaption}
\usepackage{booktabs} %
\usepackage{bm}
\usepackage{xcolor}
\usepackage{listings}

\usepackage{hyperref}

\usepackage{stmaryrd}

\usepackage[accepted]{arxiv}

\usepackage{booktabs}       %
\usepackage{amsfonts}       %
\usepackage{microtype}      %
\usepackage{url}
\usepackage{verbatim} %
\usepackage{graphicx}
\usepackage{caption} 
\usepackage{multirow}
\usepackage{xspace}
\usepackage{epsfig}
\usepackage{amsmath}
\usepackage{amsthm}
\usepackage{amssymb}
\usepackage{times}
\usepackage{xr}
\usepackage{bbm}
\usepackage{dsfont}
\usepackage{bm}
\usepackage{enumitem}
\usepackage{hyperref}       %

\newcommand{\multiline}[3][11mm]{\begin{minipage}[c][#1]{#2}#3\end{minipage}}

\newcommand{\xhdr}[1]{{\noindent\bfseries #1}.}

\newcommand{\cut}[1]{}

\newcommand{\eg}{\emph{e.g.}}

\newcommand\domain[1]{\textsc{#1}}

\newenvironment{enumerate*}%
  {\begin{enumerate}[topsep=-5pt, partopsep=0pt]%
    \setlength{\itemsep}{0pt}%
    \setlength{\parskip}{0pt}%
    \setlength{\parsep}{0pt}}%
  {\end{enumerate}}

\definecolor{codegreen}{rgb}{0,0.4,0}
\definecolor{codegray}{rgb}{0.4,0.4,0.4}
\definecolor{codepurple}{rgb}{0.5,0,0.7}
\definecolor{backcolour}{rgb}{0.96,0.96,0.94}
 
\lstdefinestyle{mystyle}{
    backgroundcolor=\color{backcolour},   
    commentstyle=\color{codegreen},
    keywordstyle=\color{magenta},
    numberstyle=\tiny\color{codegray},
    stringstyle=\color{codepurple},
    basicstyle=\ttfamily\footnotesize,
    breakatwhitespace=false,         
    breaklines=true,                 
    captionpos=b,                    
    keepspaces=true,                 
    numbers=left,                    
    numbersep=5pt,                  
    showspaces=false,                
    showstringspaces=false,
    showtabs=false,                  
    tabsize=2
}
 
\def \papertitle{Evaluating Soccer Player: from Live Camera to Deep Reinforcement Learning}
\begin{document}

\twocolumn[

\icmltitle{\papertitle}

\icmlsetsymbol{equal}{*}

\begin{icmlauthorlist}
\icmlauthor{Paul Garnier}{equal,mines}
\icmlauthor{Théophane Gregoir}{equal,SD,mines,MVA}
\end{icmlauthorlist}

\icmlaffiliation{mines}{MINES Paristech, PSL - Research University, Paris, FR}
\icmlaffiliation{SD}{SportsDynamics (see \href{https://sportsdynamics.eu/}{website})}
\icmlaffiliation{MVA}{ENS Paris-Saclay}

\icmlcorrespondingauthor{Paul Garnier}{paul.garnier@mines-paristech.fr}

\icmlkeywords{to do}

\vskip 0.3in
]

\printAffiliationsAndNotice{\icmlEqualContribution} %

\begin{abstract}
\textbf{Scientifically evaluating soccer players represents a challenging Machine Learning problem. Unfortunately, most existing answers have very opaque algorithm training procedures; relevant data are scarcely accessible and almost impossible to generate. In this paper, we will introduce a two-part solution: an open-source Player Tracking model and a new approach to evaluate these players based solely on Deep Reinforcement Learning, without human data training nor guidance. Our tracking model was trained in a supervised fashion on datasets we will also release, and our Evaluation Model relies only on simulations of virtual soccer games. Combining those two architectures allows one to evaluate Soccer Players directly from a live camera without large datasets constraints. We term our new approach Expected Discounted Goal (EDG), as it represents the number of goals a team can score or concede from a particular state. This approach leads to more meaningful results than the existing ones that are based on real-world data, and could easily be extended to other sports.}
\end{abstract}

\begin{figure}[t]
  \centering
  \includegraphics[width=0.48\textwidth]{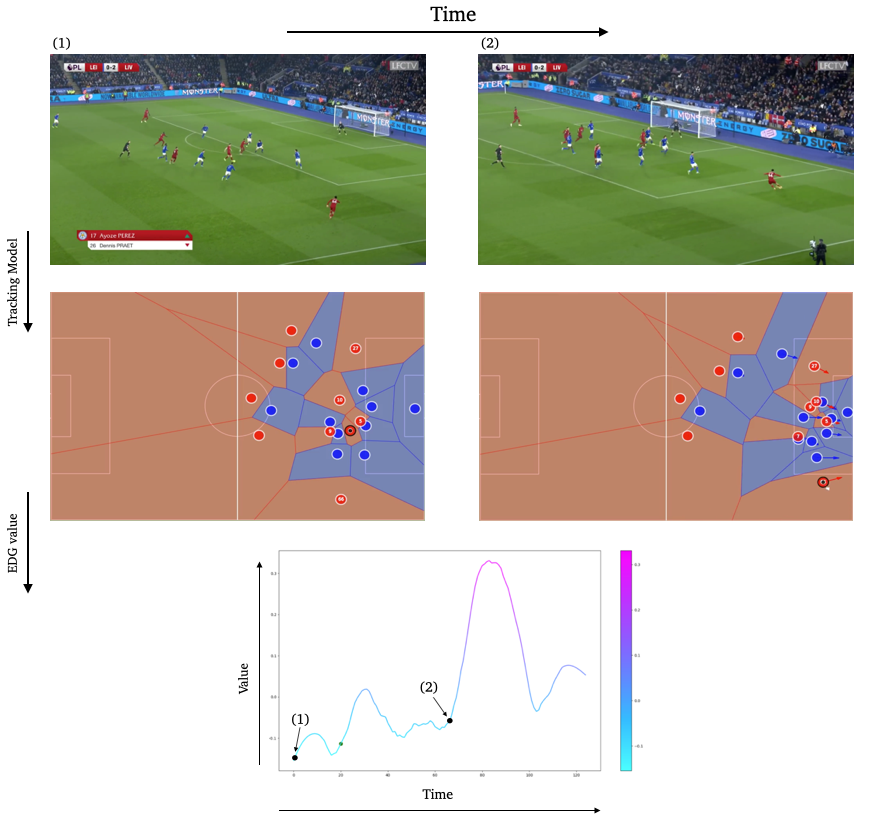}
\caption{Evolution of our Tracking and EDG model over time. The EDG model learns to capture actions with potential in a very general manner, and computes such potential with the player coordinates our Tracking model gathers from the live camera.} 
  \label{fig:teaser}
\end{figure}

\begin{figure*}[t]
  \centering
  \includegraphics[width=\textwidth]{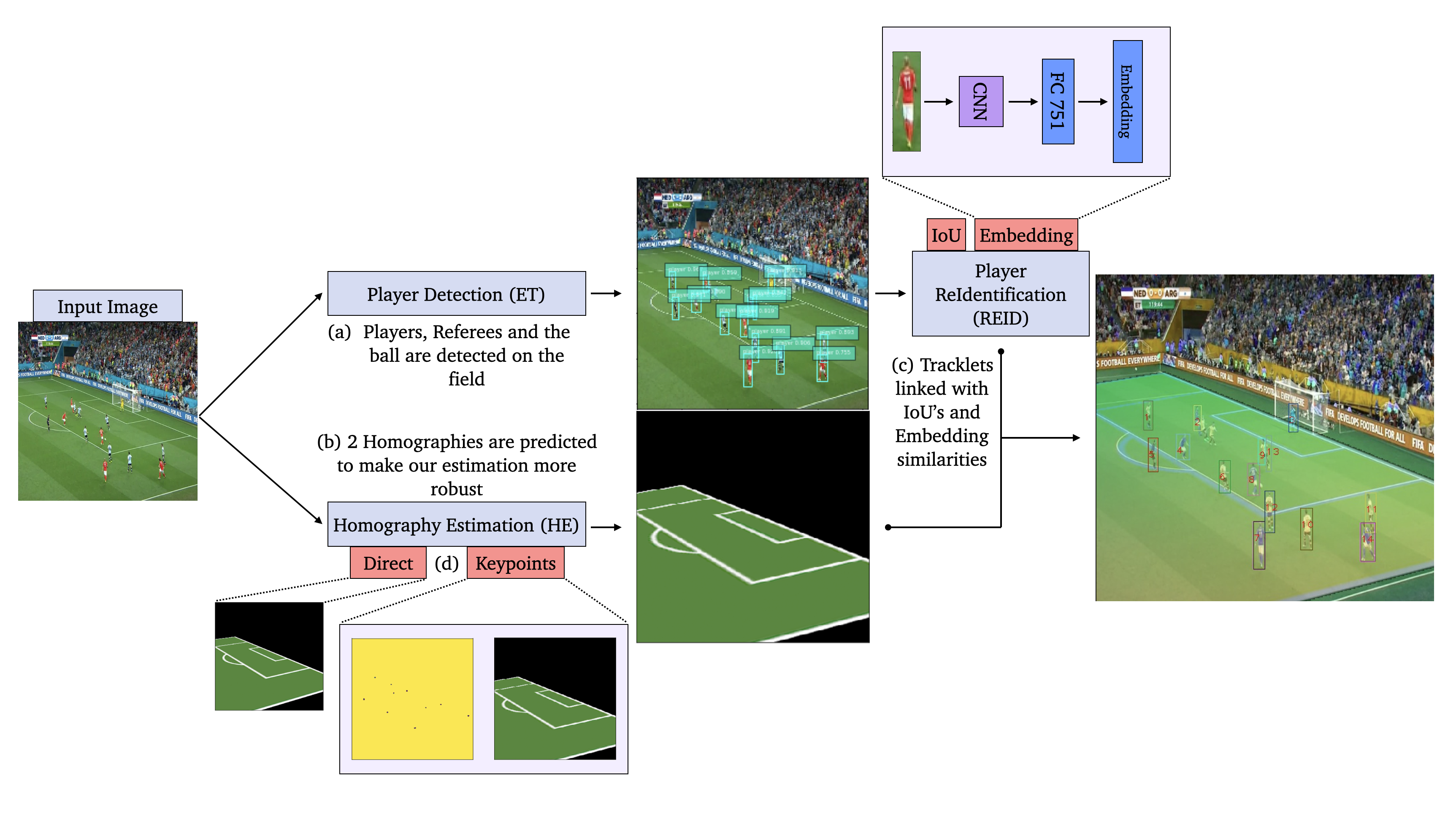}
  \caption{
  \textbf{(a)} The ET model computes the position of each entity. It then passes the coordinates to the online tracker to: (1): Compute the embedding of each player, (2): Warp each coordinate with the homography from HE.
  \textbf{(b)} The HE starts by doing a direct estimation of the homography. Available keypoints are then predicted and used to compute another estimation of the homography. Both predictions are used to remove potential outliers.
  \textbf{(c)} The REID model computes an embedding for each detected player. It also compares the IoU's of each pair of players and applies a Kalman Filter to each trajectory. 
  }
  \label{fig:schematic}
\end{figure*}

\section{Introduction}

In the 1980s, at the Hong Kong Jockey Club, Bill Benter gathered data to create a statistical prediction model for horse-racing, which made him one of the most profitable gamblers of all time \cite{Bloomberg2018}. Data became essential in most sports, impacting all aspects of their ecosystem: performance, strategy, and transfers. Advanced Sports Analytics started in the United States; franchises in Baseball, American Football, and Basketball being the first to consider data as a key for improvement and success. In charge of scouting in Baseball minor leagues for Oakland Athletics in the 1990s, Billy Beane proved that it was possible to create a cost-efficient team by finding undervalued players \cite{Lewis2003}. Using sabermetrics to value players, his 2006 Athletics team ranked 5th best of the regular season while being 24th of 30 major league teams in player salaries. Most of the popular European sports are now beginning to follow the trend, and data departments hold a more and more significant position in major clubs. Ten years after Beane’s success story, in 2016 and 2017, Liverpool FC respectively acquired Sadio Mané for £34 million and Mohamed Salah for £39 million thanks to a significant push from their director of research Ian Graham, Cambridge Ph.D. \cite{NYT2019}. These two key players led Liverpool FC to a historical 2019 Champions League victory in Madrid. Today, Sadio Mane’s and Mohamed Salah’s contracts are worth more than £100 million each.

One way of finding such an undervalued soccer player is to find alternative metrics to the traditional ones such as number of goals or passes accuracy. Here, we will present a general framework for soccer player evaluation from live camera: a tracking model processing camera's input into players' coordinates, and an \textit{Expected Discounted Goal} (EDG) model evaluating an action based on each player tracking data. The particularity of our approach comes from the idea that our EDG model is trained purely on simulation, and never on real soccer data. This makes it much easier to train, and utterly independent from large soccer datasets that are either too complicated to produce, or too expensive to acquire. 

Our EDG model generalizes well to many playstyles and many teams. It also produces similar or even more accurate results than existing models. 

Our tracking model predicts the player's coordinates and computes precise results even on difficult scenes such as sunny or covered with shadow ones. This is also the first end-to-end open-source framework for soccer player tracking, and could easily be extended to other team sports.

\section{Related Work}
\label{sec:related}

\subsection{Tracking players on a sport field} 

\xhdr{Player detection} Detecting soccer players can be a complicated task, especially in a multi-actors situation. There are multiple ways of doing this detection, \citet{Kaarthick2019} explored the usage of a HOG color based detector that gives each image a detection of the players. \citet{Neil2020} worked on a deeper model: an open-source multiperson pose estimator named \href{https://github.com/MVIG-SJTU/AlphaPose}{AlphaPose}, based on the COCO dataset. Another approach is to consider players as objects and use \emph{Object Detection Models}. \citet{Ramanathan2015} used such a model, a CNN-based multibox detector for the player detection. 

\xhdr{Field registration} One key element of any sports tracking model is how players coordinates can be represented in two dimensions. One way to do so is \emph{Sport Field Registration}, the act of finding the homography placing a camera view into a two-dimensional view, fixed across the entire game. More about Field Registration can be found in the Supplementary Materials-\ref{supp:sec:model-homography}. \citet{sharma2017} formulated the registration problem as a nearest neighbor search over a generated dictionary of edge maps and homography pairs that they made synthetically. To extract the edge data from their images, they compared 3 different approaches: Histogram of oriented gradients (HOG) features, chamfer matching, and convolution neural networks (CNN). Finally, they enhanced their findings using Markov Random Field (MRF). \citet{homayounfar2017} figured out the homography's attributes by parametrizing the problem as one of inference in an MRV which features they determine using structured SVMs. The use of synthetic data is also explored by \citet{chen2018}, which takes \citet{homayounfar2017} and \citet{sharma2017}'s work one step further by building a generative adversarial network (GAN) model. Finally, \citet{jiang2019} used a ResNet to directly predict the homography parameters iteratively, while \citet{Citraro_2020} used particular keypoints detected with a Unet from the field to achieve the same goal.

\xhdr{Entities tracking in Video}

\citet{Ramanathan2015} used a Kanade–Lucas–Tomasi feature tracker for player tracking across a game. Recent ReIdentification models go a step further: \citet{zheng2019joint} used an embedding model to extract player features, and a mixture of IoU and embedding similarities to track persons. Another approach was used by \citet{Liang2020}, with a \emph{k}-Shortest Path algorithm and an embedding model to extract players' numbers and colors.

\subsection{Valuing player actions} 

Before the revolution brought by data analysis in football, valuing players was mostly based on traditional statistics such as the number of goals, the number of assists, passes accuracy, or the number of steals. These post-game statistics can be relevant as an overview of a player’s quality, but it does not reflect how a player can impact a game thanks to his vision or his moves. Furthermore, focusing on general statistics also means ignoring the entire universe of actions in which a player took part. Thanks to the growing availability of event and tracking data, researchers started to create models using all game events and tracking data to evaluate players’ actions on and off-ball. In 2018, \citet{Spearman2018} suggested an indicator to evaluate off-ball positioning called \emph{Off-Ball Scoring Opportunities} (OBSO). Thanks to his previous fundamental works on \emph{Pitch Control} (PC), he was able to measure the positioning quality of attacking players by evaluating the danger of their position if they would receive the ball.  

The same year, using tracking data, \citet{Fernandez2018} presented two main indicators to measure the use of space and its creation during a game: \emph{Space Occupation Gain} (SOG), and \emph{Space Generation Gain} (SGG). Later on, in 2019, \citet{Fernandez2019} adapted \emph{Expected Possession Value} (EPV), a deep learning based method, from basketball to football enabling analysts to reach a goal probability estimator at each instant of a possession. By discounting EPV, researchers were able to access the impact of a single event like a key pass which would increase significantly EPV. Similarly, \citet{Decroos2019} presented their VAEP model for which they developed an entire language around event data called \emph{Soccer Player Action Description Language} (SPADL). Thanks to supervised learning, their model was trained to evaluate the impact an event could have on the scoring or conceding probability. Therefore, this model enables analysts to estimate the effect of each action from a player and then evaluate his global performance.

We start by going into the \textbf{Theoretical framework} of both our tracking model and our EDG model. In \textbf{Experimental methods}, we review the implementations and learning procedures of our models. Finally, in \textbf{Results}, we quickly present the tracking results to focus more on the insights given by our EDG model.

\section{Theoretical framework}
\label{sec:model}
\subsection{Tracking model}
\label{sec:model:tracking}
 Our EDG model needs to know the state of the game at any time $t$: we chose to do so by representing a Soccer game with a list of entities coordinates. Given a list of images, our tracking model computes the 2-dimensional trajectories of both the ball and each player. The model adopts a 3-steps method: \textsc{Entity Tracking}, \textsc{Homography Estimation}, \textsc{ReIdentification} (see Figure \ref{fig:schematic}) (ET,HE,REID). Each of these steps is represented by one or several models, and more theoretical background on the homography estimation can be found in the Supplementary Materials-\ref{supp:sec:model-homography}. The first 2 steps do not consider any temporal information as they approximate results for each image separately.  However, the last step compares each new image to the list of images that have already been processed. Doing so allows us to take into consideration each entity's movement over time and counters the mistakes the previous models might do. We quickly review here each of the 3 steps: ET, HE, REID.

\xhdr{\textsc{Entity Tracking} definition} The ET : $\mathbb{R} ^ {n \times n \times 3} \to \mathbb{R} ^ {m \times 4} \times \left[0,1\right] ^ {m} \times \left[0,1\right] ^ {m}$, takes an image as input, and predicts a list of bounding boxes associated with a class prediction (Player or Ball) and a confidence value\footnote{$m$ is defined as the number of predictions we make and is usually fixed at $100$}.

\begin{figure}[t]
  \centering
  \includegraphics[width=0.48\textwidth]{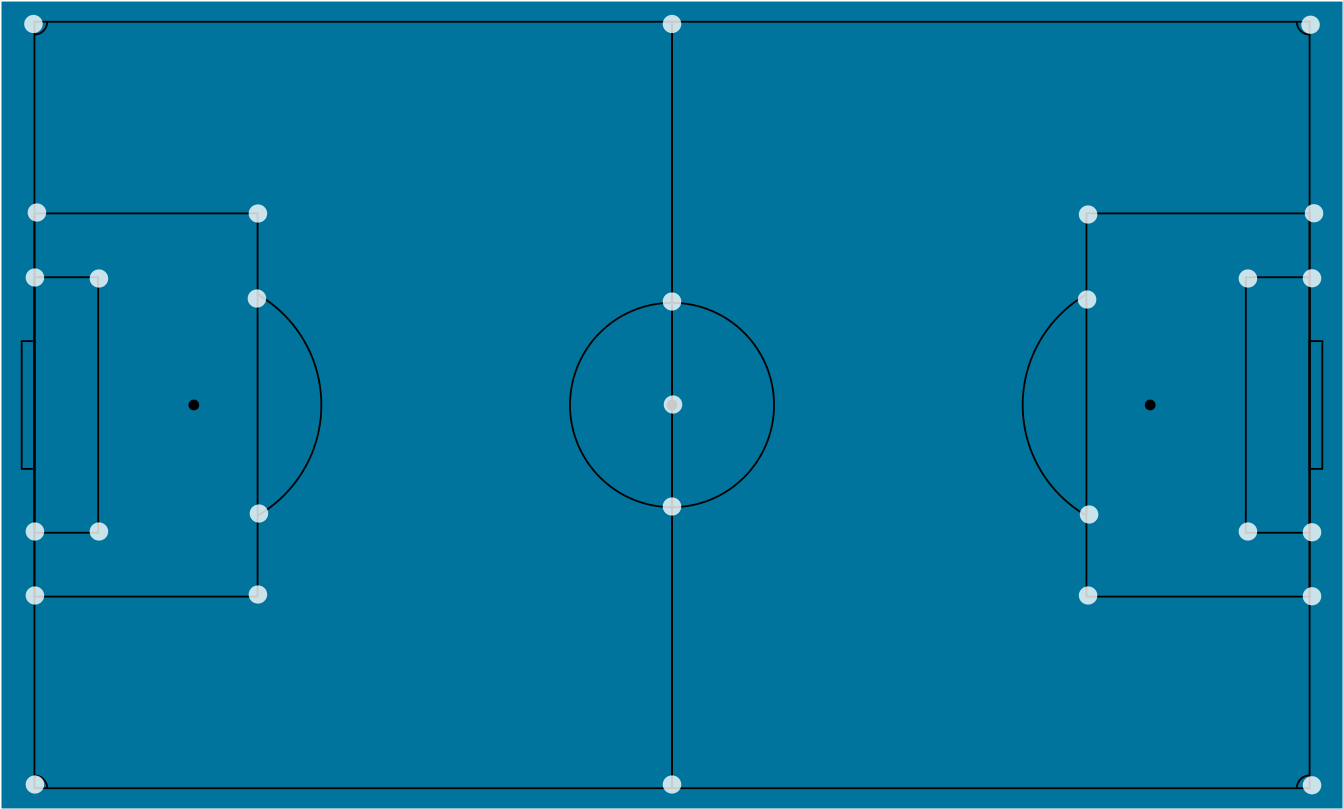}
\caption{Positions of the $28$ keypoints the HE model predicts. During training and inference, each keypoint is associated with a different mask. Such a mask is a binary matrix, with $1$s at the position of the keypoint, and $0$s elsewhere. We have $28$ masks, one per keypoint.} 
  \label{fig:field-keypoints}
\end{figure}

\xhdr{\textsc{Homography Estimation} definition} The HE is made of 2 separate models. The first one : $\mathbb{R} ^ {n \times n \times 3} \to \mathbb{R} ^ {3 \times 3}$ takes an image as input, and predicts the homography directly. The second one : $\mathbb{R} ^ {n \times n \times 3} \to \mathbb{R} ^ {p \times n \times n}$ also takes an image as input, but predicts $p$ masks, each mask representing a particular keypoint on the field (see Figure \ref{fig:field-keypoints}). The homography is computed knowing the coordinates of available keypoints on the image, by mapping them to the keypoints coordinates on a 2-dimensional field (see Supplementary Materials-\ref{supp:sec:model-homography} for more details). Using both models allows us to have better result stability, and to use one model or the other when outliers are detected.

\xhdr{\textsc{ReIdentification} definition} The REID model gathers all information from the ET model and builds the embedding model as well. The embedding model : $\mathbb{R} ^ {n \times n \times 3} \to \mathbb{R} ^ {751}$ takes the image of a detected player, and gives a meaningful embedding\footnote{751 was the initial number of classes in the Market-1501 Dataset. We kept the same number for the size of our embedding, although a smaller embedding size might lead to better results according to recent papers \cite{zheng2019joint}}. The model initializes several tracklets based on the boxes from ET in the first image. In the following ones, the model links the boxes to the existing tracklets according to: (1): their distance measured by the embedding model, (2): their distance measured by boxes IoU's. A Kalman Filter \cite{Kalman1960ANA} is also applied to predict the location of the tracklets in the current image and removes the ones that are too far from the linked detection. The model also updates the embedding of each player over time with another Kalman Filter.

\subsection{EDG model}
\label{sec:model:theory}

We assume $s_t \in \mathcal{S}$ is the state of the game at time $t$. It may be the positions of each player and the ball for example. Given an action $a \in \mathcal{A}$ (\eg a pass, a shot, more details in the Supplementary Materials-\ref{supp:sec:datasets}), and a state $s' \in \mathcal{S}$, we note $\mathbb{P} \colon \mathcal{S} \times \mathcal{A} \times \mathcal{S} \to [0,1]$ the probability $\mathbb{P} (s' \vert s, a)$ of getting to state $s'$ from $s$ following action $a$. 
Applying actions over $K$ time steps yields a trajectory of states and actions, $\tau ^{t_{0:K}} = \big(s_{t_0},a_{t_0}, ... ,s_{t_K},a_{t_K} \big)$. We denote $r_t$ the reward given going from $s_t$ to $s_{t+1}$ (\eg $+1$ if the team scores a goal). More importantly, the cumulative discounted reward along $\tau ^{t_{0:K}}$ is defined as:

$$R(\tau ^{t_{0:K}}) = \sum_{n=0}^{K} \gamma^{n} r_{t_n}$$

where $\gamma \in \left[0,1\right]$ is a discount factor, smoothing the impact of temporally distant rewards.

A policy, $\pi _{\theta}$, chooses the action at any given state, whose parameters, $\theta$, can be optimized for some training objectives (such as maximizing $R$). Here, a good policy would be a policy representing the team we want to analyze in the right manner. The \textit{Expected Discounted Goal} (EDG), or more generally, the state value function, is defined as:

$$V^\pi (s) = \underset{\tau \sim \pi}{\mathbb{E}} \big[ R(\tau) \vert s \big]$$

It represents the discounted expected number of goals the team will score (or concede) from a particular state. To build such a good policy, one can define an objective function based on the cumulative discounted reward:

$$J(\theta) = \underset{\tau \sim \pi_\theta}{\mathbb{E}} \big[ R(\tau) \big]$$

and seek the optimal parametrization $\theta ^*$ that maximize $J(\theta)$:

$$\theta^* = \arg \max_\theta \mathbb{E} \big[ R(\tau) \big]$$

To that end, we can compute the gradient of such cost function\footnote{Using a log probability trick, we can show that we have the following equality: $\nabla_\theta J(\theta) = \underset{\tau \sim \pi_\theta}{\mathbb{E}} \left[ \sum_{t=0}^T \nabla_\theta \log \left( \pi_\theta (a_t \vert s_t) \right) R(\tau) \right]$} $\nabla_\theta J(\theta)$ to update our parameters with $\theta \leftarrow \theta + \lambda \nabla_\theta J(\theta)$. In our case, the evaluation of $V^\pi$ and $\pi_\theta$ is done using Neural Networks, and $\theta$ represents the weights of such networks (more details on Neural Networks can be found in \citet{Goodfellow2017}). At inference, our model will take the state of the game as input, and will output the estimation of the EDG. A more advanced view of Reinforcement Learning can be found in \citet{SuttonDRL}

\section{Experimental methods}
\label{sec:experiments}

All of the models were implemented using standard deep learning libraries and the usual image processing library. Models, code, and datasets will be released at \href{https://github.com/DonsetPG/narya}{https://github.com/DonsetPG/narya}. 

\subsection{Tracking implementation details}
\label{sec:experiment:tracking}

\xhdr{\textsc{Entity Tracking} details} The ET model is based on a Single Shot MultiBox Detector (SSD) \cite{Liu_2016}. The model takes images of shape $(512,512)$, and makes prediction for 2 classes: players (including referees) and the ball. We used an implementation from GluonCV \cite{gluoncvnlp2020}.

\xhdr{\textsc{Homography Estimation} details} The first model (direct estimation) is based on a Resnet-18 architecture \cite{he2015deep}. It takes images of shape $(280,280)$ and randomly crops images of shape $(256,256)$ during training. The model doesn't directly estimate the homography, but rather the coordinates of 4 control points, in the same manner as \cite{jiang2019}. More details can be found in Supplementary Materials-\ref{supp:sec:model-homography}. The model is implemented with Keras \cite{chollet2015keras}. The second model is based on an EfficientNetb-3 backbone \cite{tan2019efficientnet} on top of a Feature Pyramid Networks (FPN) \cite{lin2016feature} architecture to predict the mask of each keypoint. We implemented our model using Segmentation Models \cite{Yakubovskiy:2019}.

\xhdr{\textsc{ReIdentification} details} The embedding model is based on a Resnet-50 architecture \cite{he2015deep} to produce embeddings of size $751$. The model was produced for supervised Re-Identification \cite{zheng2019joint} and implemented on Torch \cite{paszke2017automatic}.

\subsection{DRL implementations details}
\label{sec:experiment:drl}

The EDG model is based on a Deep Reinforcement Learning environment \cite{kurach2019} built by the Google Brain Team and based on a gym environment \cite{openai}.

\xhdr{States, actions, and reward representations} Each state $s$ is represented by a \textit{Super Mini Map} (SMM), stacked 4 times to represent the last 4 positions of each entity. More precisely, $s \in \mathbb{R} ^ {72 \times 96 \times 16}$, where each consecutive 4 matrices encode the positions of the home team, the away team, the ball, and the active player (leading to $ 4 \times 4 = 16 $ for third dimension). The encoding is binary in $\{0,255\}$, representing whether an entity is at the given coordinates or not.

The actions $a$ include movements in 8 directions, different kicks (\eg \; short and long passes, shooting), and high passes. Each action is sticky, meaning that once executed, a moving or sprinting action will not stop until explicitly told to do so. The reward $r \in \{-1,0,1\}$ represents if a goal is conceded, not happening or scored.

\begin{figure}[t]
  \centering
  \includegraphics[width=0.48\textwidth]{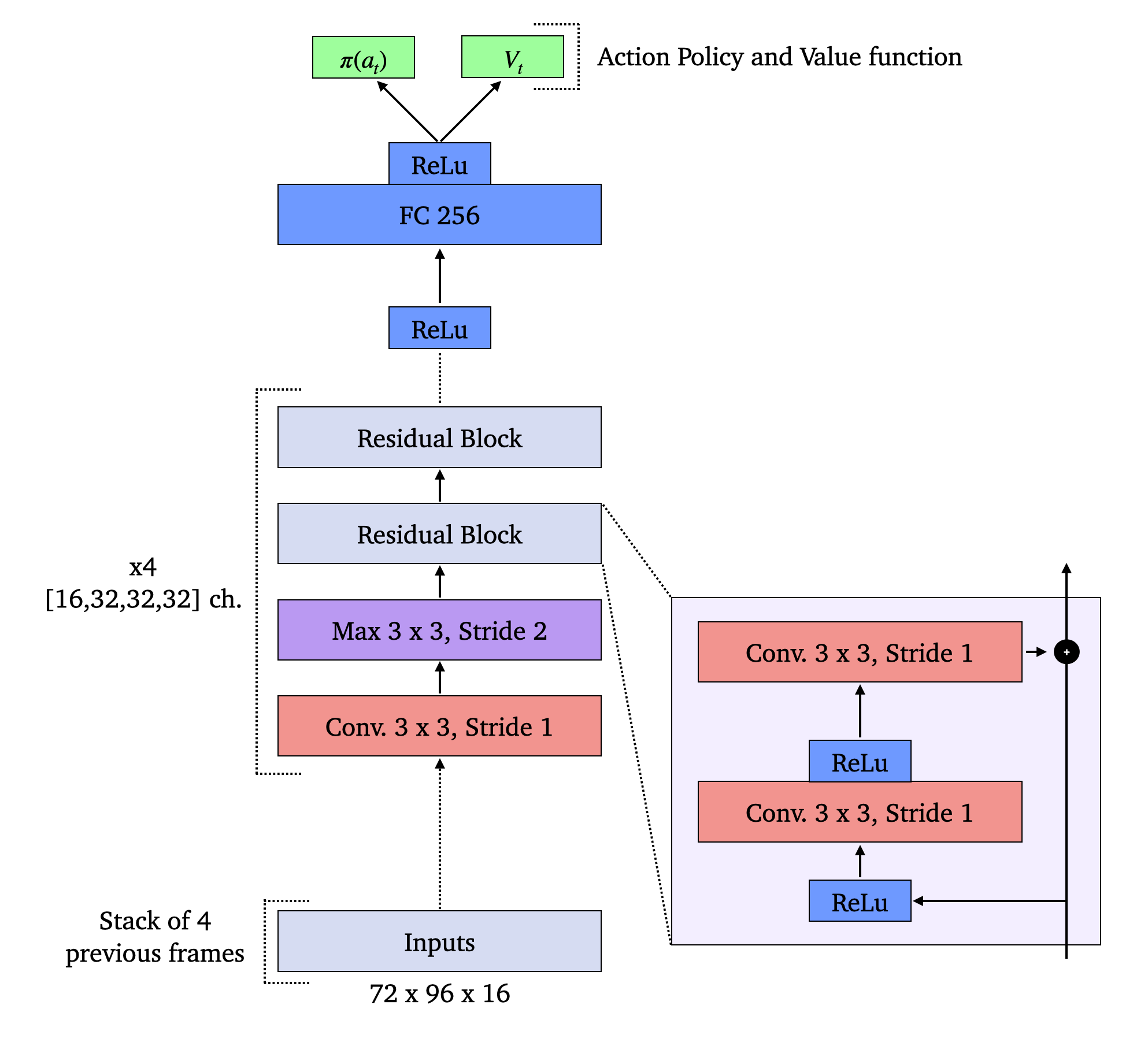}
\caption{Architecture used for our agent.} 
  \label{fig:agent-archi}
\end{figure}

\xhdr{EDG model details} The DRL agent, capturing the policy and the EDG is represented by a PPO \cite{schulman2017proximal} algorithm, with an Impala policy \cite{espeholt2018impala}. The architecture is available Figure-\ref{fig:agent-archi}.

\subsection{Training}
\label{sec:experiment:training}

\subsubsection{Tracking models}

\xhdr{Datasets} We built 3 datasets: a tracking dataset, a homography dataset, and a keypoint dataset. They all consist of the same images from the 2014 World Cup, the Premier League, the Ligue 1, and the Liga. They typically contain $450$ train images, $50$ validation images, and $50$ test images. The homography dataset was based on an existing 2014 World Cup dataset \cite{homayounfar2017}. We extended it to the rest of our images. More details on the datasets can be found in the Supplementary Materials-\ref{supp:sec:datasets}.

\xhdr{Data augmentation} The models were trained on a Tesla-P100 GPU, and each model takes a few hours to train. In each training we used data augmentation : (1): Use the entire original input image, (2): Randomly flip the image horizontally, (3): Randomly add Gaussian Noise and shadows on the field, (4): Randomly change the brightness, the contrast, and the saturation, (5): Add motion blur.

\xhdr{Optimization procedures} Both of the homography models were trained for $200$ epochs, with a batch size of $32$ and a learning rate of $10^{-4}$. We used an exponential learning rate decay down to $10^{-8}$. The tracking model was trained for $100$ epochs, with a batch size of $2$ and a learning rate of $10^{-3}$. The embedding model was pretrained on the Market-1501 dataset \cite{zheng2015scalable}. Each model was evaluated each $5$ training epochs, and we stopped training when we observed a negligible decrease in the loss function. Each model was trained using an Adam optimizer \cite{kingma2014adam}.

\subsubsection{DRL Agent} 

\xhdr{Selfplay} We started with a pretrained Agent from Google, trained from 50M steps against an easy bot. We then kept training it for 50M steps against a medium bot. Finally, we trained the agent another 50M against the last version of itself twice. While previous results \cite{kurach2019} were mitigated about self-play, we believe that given the significant advantage our agent had against a medium bot, it was the right decision to improve its capability to find actions with potentials. We compare the results of agents with and without selfplay Figure \ref{fig:res-value} \textbf{(right)} and in the Supplementary Materials-\ref{supp:sec:drl-supp-results}.

\xhdr{Optimization procedures} The agent was trained on $8$ parallel environments with a learning rate of $.000343$ and a discount factor $\gamma$ of $0.993$. We use $8$ minibatches per epoch and update our agent $4$ times per epoch. Finally, the agent is updated with a batch size of $1024$, using an Adam optimizer again. More details on the hyperparameters can be found in the Supplementary Materials-\ref{supp:sec:drl-hyp} Figure \ref{supp:fig:us-vs-easy}.

\subsection{Evaluation}
\label{sec:experiment:evaluation}

The homography estimation models use the IoU between the warped field with our estimation and ground truth homography as the primary metric. We used the mean Average Precision (mAP) for the tracking model for bounding boxes with an IoU greater than $0.5$. Finally, we did not use any metric for the ReIdentification model, as the model we used was not finetuned on a Soccer player dataset. Mode details on the metrics, results, and hyperparameters can be found in the Supplementary Materials-\ref{supp:sec:results}.

We report the result of our DRL agent both with the Average Goal Difference against its opponent and by comparing the EDG to existing frameworks on the same actions. This allows both to grasp the policy's level and how close it reacts to policy based on real players. 

\section{Results}
\label{sec:results}

Our main findings are that the EDG model can have a relevant representation of the potential of an action. It can even detect events that previous frameworks could not, even though the agent never trained on real game data. Furthermore, the tracking model can produce reliable coordinates for short plays from only one camera. This model allows us to produce the EDG of any given action quickly. We start by quickly reviewing the results of our Tracking model in Section \ref{sec:results:tracking}. In Section \ref{sec:results:edg} and \ref{sec:results:comparison}, we present the results of the EDG model and compare it to existing value frameworks.

We kept the same agent and the same tracking models throughout the results to challenge their robustness to different scenarios or teams.

\subsection{Generating players' coordinates}
\label{sec:results:tracking}

\begin{figure*}[t!]
  \centering
  \includegraphics[width=\textwidth]{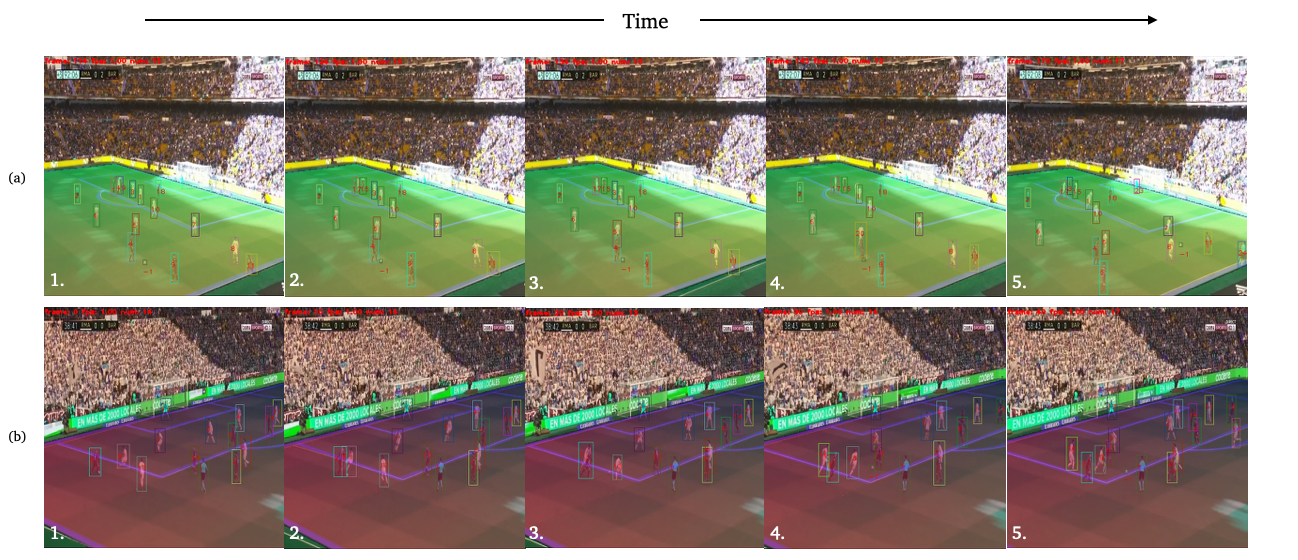}
  \caption{We can track almost every player and the ball, even in very different settings. We present here some examples of tracking failures and success. In sequence (a), the REID model fails to identify player 19 between frame 1. and 2., letting the same player having a new id (15). Between frames 3., 4. and 5., the boxes of players 4 and 5 merge into id 20. When the merge is undone, the model can retrieve the right id for each player \newline 
  In sequence (b), the same phenomenon appears frame 2., 3. and 4. but the model fails to recognize player 8 and gives him a new id (16). Another issue with our model is when a player runs out of the camera's range and returns later. This almost always leads to a change of id, as seen in frame 5. with player 17. However, this can easily be managed by merging the two ids trajectories.}
  \label{fig:res-tracking}
\end{figure*}

We found that each component of the Tracking Model generalizes well to many scenarios. We start by reviewing the results for each component and leave discussion about some architectural choices to the Supplementary Materials. 

\begin{table}[t!]
\begin{center}
\begin{tabular}[p]{lcc}

\toprule

            \textbf{Method} & \textbf{IoU - Mean} & \textbf{IoU - median} \\

\hline 

OURS & 0.908 & 0.921   \\
OURS, keypoints & 0.892 & 0.923   \\ 
OURS, direct estimation & 0.891 & 0.906   \\

\hline 

Keypoints, with Players$^a$ & 0.939 & 0.955   \\
Keypoints, w/o Players$^a$ & 0.905 & 0.918   \\

\hline 

Synthetic Dictionary$^b$ & 0.914 & 0.927   \\

\hline 

Learned Errors$^c$ & 0.898 & 0.929   \\

\hline 

Branch and Bound$^d$ & 0.83 & -   \\

\hline 

PoseNet$^e$ & 0.528 & 0.559   \\

\hline 

SIFT$^f$ & 0.170 & 0.011   \\

\hline 

\bottomrule
\end{tabular}
\caption{\label{tab:res-homo}Quantitative results of the Homography estimator models on the 2014 World Cup dataset. Even though \cite{Citraro_2020} obtain better results with their method when using the positions of players as well, we believe that our results where good enough for our purpose, which was to provide reliable tracking data to our EDG model. \newline 
\footnotesize{$^a$ \cite{Citraro_2020}, $^b$ \cite{sharma2017}, $^c$ \cite{jiang2019},  $^d$ \cite{homayounfar2017},  $^e$ \cite{kendall2015},  $^f$ \cite{Lowe2004}}}
\end{center}
\end{table}

\begin{table}[t!]
\begin{center}
\begin{tabular}[p]{l|cc|c}

\toprule

\textbf{Backbone} & \textbf{Player - AP} & \textbf{Ball - AP} & \textbf{mAP} \\

\hline 

VGG16 & 89.3 & 39.5 & 64.4   \\
Resnet50 & 89.9 & 59.3 & 74.6  \\

\bottomrule
\end{tabular}
\caption{\label{tab:res-tracking-gluon}Quantitative results of the player Tracking model, with a SSD model. More details about the results for less impactfull choices can be found in the Supplementary Materials-\ref{supp:sec:results}}
\end{center}
\end{table}

\begin{table}[t!]
\begin{center}
\begin{tabular}[p]{l|c|cc}

\toprule

\textbf{Architecture} & \textbf{Backbone} & \textbf{IoU} & \textbf{F1-Score} \\

\hline 

Unet & EfficientNetb-3 & 0.817 & 0.831\\

\hline 

LinkNet & EfficientNetb-3 & 0.813 & 0.828\\

\hline 

FPN & EfficientNetb-3 & 0.882 & 0.91 \\
    & Resnet18 & 0.862 & 0.885 \\
    & Resnet50 & 0.841 & 0.863 \\

\bottomrule
\end{tabular}
\caption{\label{tab:res-masks}Quantitative results for the keypoints masks generations. More details about the results for other backbones can be found Figure \ref{supp:fig:keypoints-ablation}}
\end{center}
\end{table}

\xhdr{\textsc{Entity Tracking} results} The main results are available in Table \ref{tab:res-tracking-gluon}. Overall, the Resnet50 backbone yields much better results than the VGG16. The Average Precision for the player class is about $90\%$, where the mistakes often come from stacks of players that are hard to distinguish even for the human eye. Because of its relatively small size and its frequent occlusion by players, the ball Average Precision does not get higher than $60\%$. 

\xhdr{\textsc{Homography Estimation} results} The results and comparison against existing methods are available Table \ref{tab:res-homo}. We find similar results to \cite{Citraro_2020} for our Keypoints based method and slightly better results than the direct estimation implementation from \cite{jiang2019}. This can be explained by the higher number of dense layers added at the end of our Resnet18 architecture. Our combination of 2 very different strategies allows the complete model to react better to outliers and bypass the condition of 4 non-collinear points needed by the keypoints based model. Producing high-quality estimations of the homography is a hard task with many consequences. If an estimation is incorrect, players' coordinates will be as bad and produce noisy or false trajectories. For this reason, it is useful to manually check some estimations by hand during the tracking process, and remove them if they are too far from the truth. The more detailed results of the keypoints mask prediction are available in Table \ref{tab:res-masks}.

\xhdr{\textsc{ReIdentification} results} Since we have no dataset of soccer player re-Identification, it is impossible to evaluate our embedding model. However, we can denote the number of operations we had to do manually after a tracking is produced. On average, we had to delete $3.5$ trajectories per action (one trajectory to remove every $60.57$ frame, or $3$ seconds, on average). This number includes the referees that we also detect in our dataset and model. We merge $5$ trajectories per action on average (\eg when a player is not recognized and gets a new id). Finally, we manually add $3.75$ missing coordinates on average, mostly when we do not find the ball at a critical moment (\eg \; the beginning of a pass or a shot).

We present the results of our entire model Figure \ref{fig:res-tracking} for $2$ different actions, and give a more in-depth analysis of when our model successfully tracks each player even when the tracking model fails to detect them, but also highlight particular cases of Re-Identification failure. We discuss in the Supplementary Materials-\ref{supp:sec:results} the rest of architectural choices. We also display more results for each model in Figure \ref{fig:full-results}.

\subsection{Valuing expected discounted goals}
\label{sec:results:edg}

\begin{figure*}[!t]
  \centering
  \includegraphics[width=\textwidth]{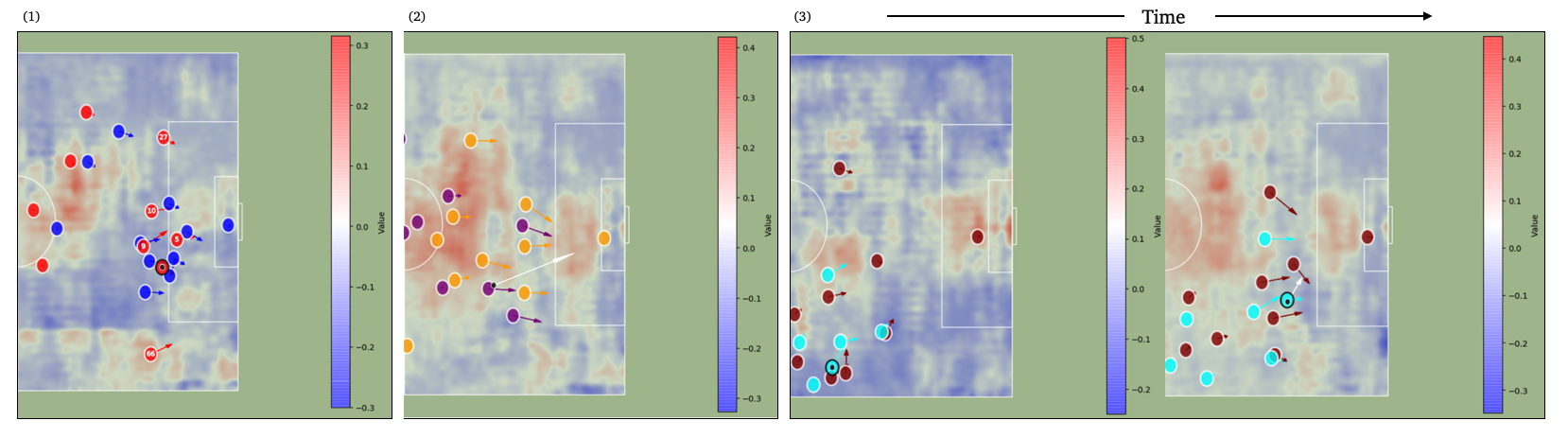}
  \caption{EDG mapping based on the position of the ball. In (1), passing the ball to the midfield would release pressure and increase the EDG, but lose momentum. A forward pass inside the box or to the right-wing would also increase the EDG and keep the current momentum. In (2), a forward pass to the right-wing would have less impact as one inside the box, given the current players' velocities. We present in (3) an evolution of the EDG overtime. In the first frame, a pass to the middle would help release the high pressure. The team prefers to stay on the right side and to create a momentum towards the goal. The EDG evolves then to increase on the wing. The EDG does not take into account the different physical capabilities of each player, thus underestimate the potential the team attack's speed.}
  \label{fig:res-edg}
\end{figure*}

\begin{figure*}[!t]
  \begin{tabular}[p]{l|r}
  \begin{minipage}{4.04in}
  \includegraphics[trim=2.5mm 0 0 0,width=\textwidth]{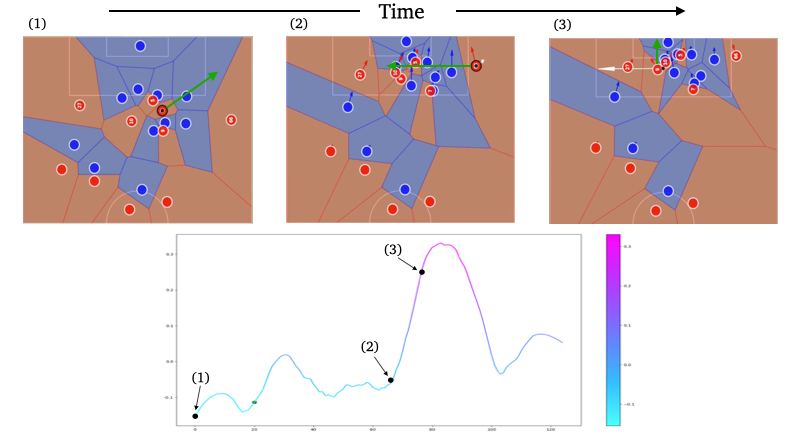}%
  \end{minipage}
& {\begin{minipage}{2.49in}
  \includegraphics[width=1.\textwidth]{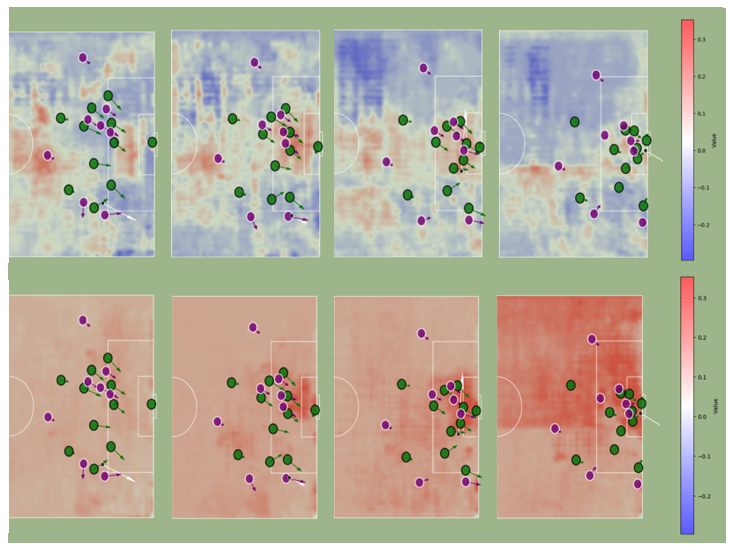} \\ 
  \end{minipage}}
\end{tabular}

  \caption{\textbf{(left)} We present here the EDG over time alongside the tracking of players. The green arrows represent the action taken in the next frame by the player. Between frame (1) and (2), the EDG increases with the long forward pass to the right-wing. It keeps increasing frame (2) to (3) with a decisive pass inside the box, before decreasing due to the very high pressure. It increases again with a shot. \\
  \textbf{(right)} Comparison between our agent EDG map and an agent trained for 50M steps against an easy bot. The easy agent attribute on average the same potential to every location on the field. It shows how training against a more difficult agent and self-play helped our agent learn real soccer players' behavior.}
  \label{fig:res-value}%
\end{figure*}

Our Agent imparts a satisfactory understanding of soccer and soccer dynamics, even while training on simulations. First, Table \ref{res:table:rewards} shows the Average Goal Difference between our Agent and its opponent at the end of each training step ($50$M updates): not only does the Agent learn to beat easy and medium bots, but it also learns more complex dynamics while training against itself. The major impact of this selfplay component can be seen Figure \ref{fig:res-value} \textbf{(right)}, and in the Supplementary Materials Figure \ref{supp:fig:us-vs-easy}. We displayed on these maps the EDG estimated by our Agent if the ball was located at this location. After the first 2 steps of training, the Agent attributes on average the same EDG to each location of the field. After selfplay, a more realistic and insightful potential is discovered.

This potential can be seen Figure \ref{fig:res-edg} for 4 different games. While the front of the center circle and penalty sport both get a high EDG (for releasing the pressure and a high probability of goal respectively), the Agent also accurately captures the most dangerous zones. For example, in both Figure \ref{fig:res-edg} (1) and (2), a forward pass to the right-wing would bring a lot of value to the action. Since the Agent inputs take into consideration the four last frames, it can consider each entity's velocity. 

Our model can also estimate the EDG of an action over time. We present  \ref{fig:res-value} \textbf{(left)} the evolution of the EDG during an action, with the player tracking as well. While our Agent can capture the potential of an action, we believe that 2 components make the EDG extremely valuable for soccer analysis:

\xhdr{The discount factor, $\gamma$} We chose a discount factor of $0.993$, meaning that if a goal takes place 100 frames later, it brings a value of $0.993^{100} \approx 0.5$. It reduces the bias from which a single goal would bring immense value to each previous action, even ones that are far away. This allows linking each action to the goal they might lead to realistically, without any arbitrary tricks\footnote{Such as completely splitting the relations between passes and shots, for example, or attributing the same reward to each action that led to a goal, or only to the last 10.}. The value of a goal is, therefore, more uniformly distributed over actions.

\xhdr{The expected value, $\underset{\tau \sim \pi_\theta}{\mathbb{E}}$} First, the Expected Value allows the EDG to get closer to the concept of \emph{almost-goal}, which means that we want to remove any randomness in soccer and consider equal a goal or a shot striking the goalposts for example. Equivalently, we don't want to consider too many outliers, such as lucky goals. This has two effects: a better estimation of the real potential of an action, and better detection of undervalued (resp. overvalued) actions and players. Most importantly, it forces analysis to focus on what leads to a goal, not on the shot itself.

\begin{table}[h]
\begin{center}
\begin{tabular}[p]{l|c|c}

\toprule

\textbf{Training Step} & \textbf{Opponent} & \multiline{25mm}{\centering\textbf{Average \\ Goal Difference}} \\
\hline 

Step 1 & Easy Bot & 8.14  \\
Step 2 & Medium Bot & 4.7 \\
Step 3 & end of Step 2 & 5.3\\
Step 4 & end of Step 3 & 4.8\\

\bottomrule
\end{tabular}
\caption{\label{res:table:rewards}Average Goal Differences at the end of the training between our Agent and the opponent.}
\end{center}
\end{table}

\subsection{Comparison with existing framework}
\label{sec:results:comparison}

\begin{figure}[!t]
  \centering
  \includegraphics[width=0.48\textwidth]{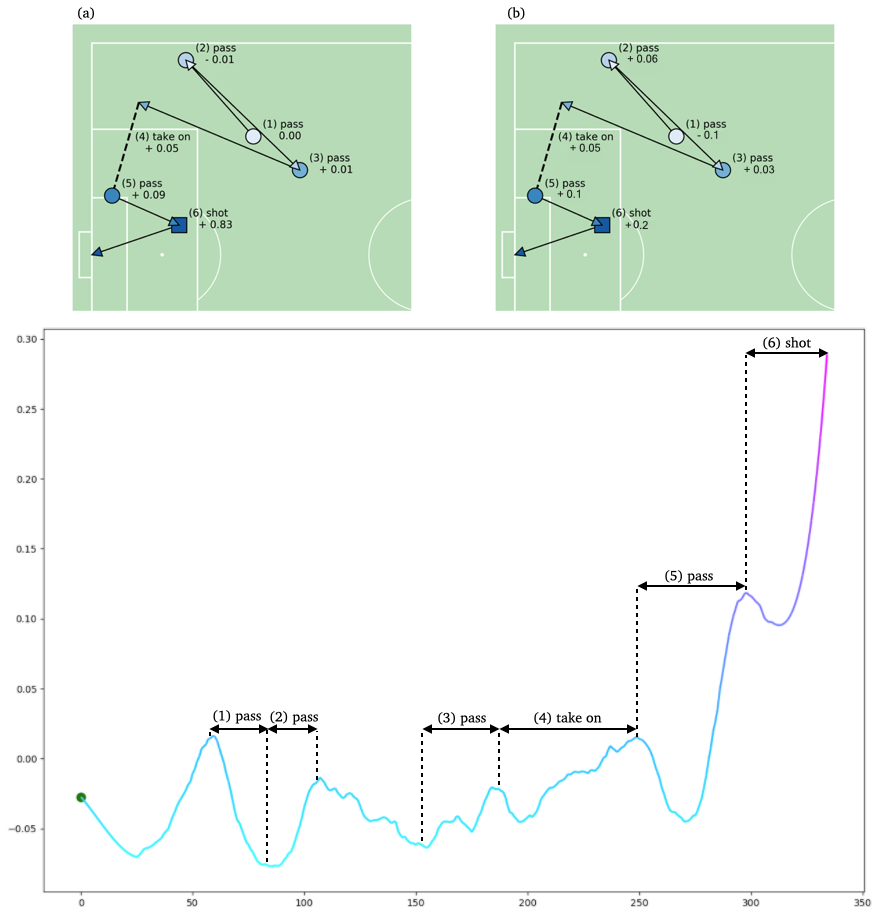}
\caption{We compare here on the same scenario, VAEP \cite{Decroos2019} (a) and our EDG (b). Our EDG is negative on pass (1), as we take into account the 2 close surrounding opponents to the receiver. The next 2 passes are highly important, and get more values in our framework as they allow to: (2) release pressure, (3) add momentum and danger close to the goal. The next two actions get overall the same value as the VAEP. However, the shot gets a much smaller value within our framework, as our model discounts the goal and attributes more uniformly the potential of an action.} 
  \label{fig:comp-vaep}
\end{figure}

\begin{figure*}[!t]
  \centering
  \includegraphics[width=\textwidth]{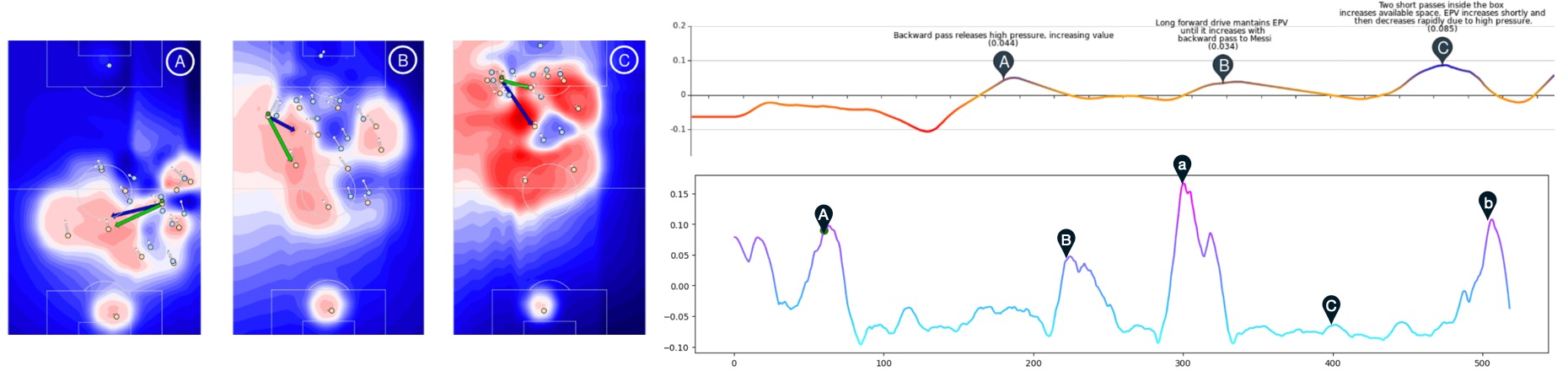}
  \caption{We compare here in the same scenario the EPV \cite{Fernandez2019} and our EDG. We pick up the same increase in value for action A and B, while our model doesn't value as much action C. On the other hand, our model values a lot action a, which is a long forward pass to the left-wing, and a strong situation that could lead crossing the ball to a teammate. We also kept track of the end of the action (situation b) with a great pass followed by a caught header.}
  \label{fig:comp-VAEP}
\end{figure*}

One way to ensure our Agent produces good analysis is to compare it to other algorithms whose purpose is to evaluate soccer players. We focus on 2 different models: the Expected Possession Value (EPV) from \citet{Fernandez2019}, and Valuing Actions by Estimating
Probabilities (VAEP) from \citet{Decroos2019}. We compare the three approaches to the scenario they used on their paper, as we do not have access to their training procedures or their data. 

Overall, we find that our approach, while never trained on data extracted from a real game, find on average the same value as the other 2 approaches. The EPV and our Agent always agree on the positive or negative impact of an action, and when the surrounding is not too impactful, the VAEP and our Agent also agree on the value of an action\footnote{Since the VAEP doesn't take into consideration if opponents are close to a player for example}.

It does, however, find more precise and meaningful values for more subtle actions. This is a powerful result and proof that our Agent learnt realistic soccer behaviors in a simulation, and can understand and process information at various levels: \eg action a player can perform, pressing on a player, players' velocities. For example, a forward pass would get a good value from the VAEP algorithm, while our Agent could give him a weak value because opponents are close, or if a pass is risky because of the field limits. Overall, we find that the value of a goal is much more uniformly distributed in the previous actions with our framework. For example, an assist and its related goal would get values distribution of 10\% and 90\% in the other frameworks, while our Agent would give a distribution of 35\% and 65\%.

\section{Conclusion}
\label{sec:conclusion}

We presented a generic framework for soccer player tracking and evaluation based on machine learning and Deep Reinforcement Learning. Our experiments show that we can produce trajectories of players only from a single camera and that our EDG framework captures more insightful potential than previous models, while being based on simulations. We find that both the tracking model and EDG agent are the first open-sourced one while being more accurate than previous approaches. 

Regarding the tracking models, our model could easily be extended to other sports, with appropriate datasets. Some improvements may be obtained by building a soccer ReIdentification datasets or building Unsupervised Approach. 

Our DRL Agent would benefit a lot from being replaced by a multi-agent algorithm, where more than one player are being controlled at once. Improvements over the soccer environment would also help a lot: bringing more physical statistics for players or particular teams, for example. While we focus on pure simulation-based training, finetuning the Agent on Expert Data from real games could help to bridge the gap with a more realistic playstyle. 

Finally, this work aims to give a broader audience access to soccer tracking data and analysis tools. We also hope that our datasets will help to build more performant models over time.

\section*{Acknowledgements}

We thank Arthur Verrez and Gauthier Guinet for valuable feedback on the work and manuscript, and we thank Matthias Cremieux for advice on object detection models. We also thank \href{https://twitter.com/lastrowview}{Last Row} for the first tracking data we used to test our agent.

\bibliography{main}
\bibliographystyle{icml2020}

\cut{
\appendix
\section{Do \emph{not} have an appendix here}

\textbf{\emph{Do not put content after the references.}}
Put anything that you might normally include after the references in a separate
supplementary file.

We recommend that you build supplementary material in a separate document.
If you must create one PDF and cut it up, please be careful to use a tool that
doesn't alter the margins, and that doesn't aggressively rewrite the PDF file.
pdftk usually works fine. 

\textbf{Please do not use Apple's preview to cut off supplementary material.} In
previous years it has altered margins, and created headaches at the camera-ready
stage. 
}

\appendix

\counterwithin{figure}{section}
\counterwithin{table}{section}
\counterwithin{algorithm}{section}
\onecolumn
\icmltitle{Supplementary Material: \papertitle}

\section{Supplementary Homography model details}
\label{supp:sec:model-homography}

\begin{figure*}[h]
  \centering
  \includegraphics[width=\textwidth]{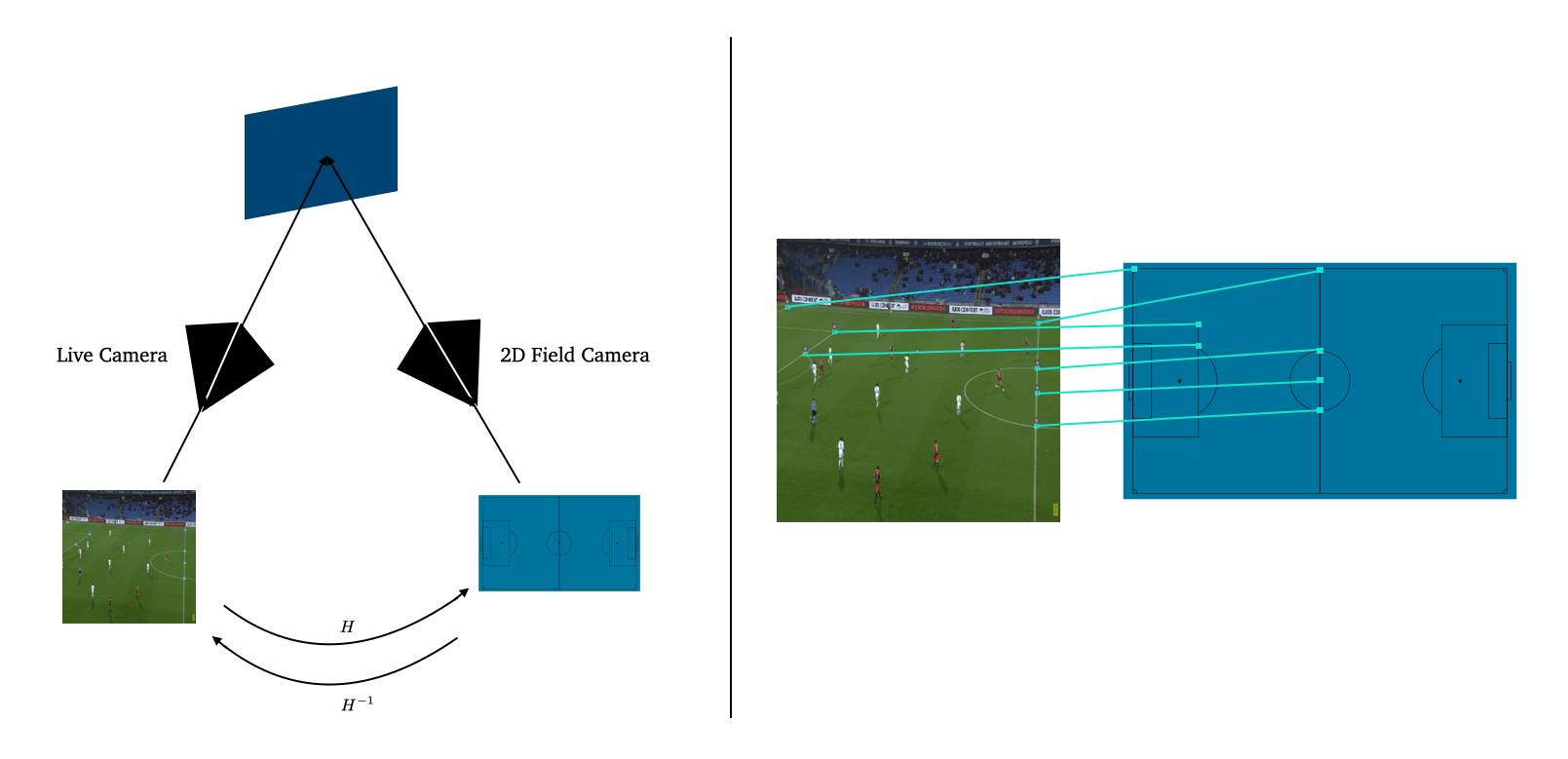}
  \caption{\textbf{(left)} Example of a planar surface (our soccer field) viewed by two camera positions: the 2D field camera and the Live camera. Finding the homography between them allows to produce 2D coordinates for each player. \textbf{(right)} Example of points correspondences between the 2D Soccer Field Camera and the Live Camera.}
  \label{supp:fig:eg-homography}
\end{figure*}

We assume 2 sets of points $(x_1,y_1)$ and $(x_2,y_2)$ both in $\mathbb{R}^2$, and define $\mathbf{X_i}$ as $[x_i,y_i,1]^{\top}$. We define the planar homography $\mathbf{H}$ that relates the transformation between the 2 planes generated by $\mathbf{X_1}$ and $\mathbf{X_2}$ as : 

\begin{align*}
    \mathbf{X_2} = \mathbf{H}\mathbf{X_1} =  \begin{bmatrix} h_{11} & h_{12} & h_{13} \\ h_{21} & h_{22} & h_{23} \\ h_{31} & h_{32} & h_{33} \end{bmatrix} \mathbf{X_1}
\end{align*}

where we assume $h_{33} = 1$ to normalize $\mathbf{H}$ and since $\mathbf{H}$ only has $8$ degrees of freedom as it estimates only up to a scale factor. An example of such homography is available Figure \ref{supp:fig:eg-homography} \textbf{(left)}. The equation above yields the following 2 equations: 

\begin{align*}
    x_2 = \frac{h_{11}x_1 + h_{12}y_1 + h_{13}}{h_{31}x_1 + h_{32}y_1 + 1} 
    \\
    y_2 = \frac{h_{21}x_1 + h_{22}y_1 + h_{23}}{h_{31}x_1 + h_{32}y_1 + 1} 
\end{align*}

that we can rewrite as : 

\begin{align*}
    x_2 = h_{11}x_1 + h_{12}y_1 + h_{13} - h_{31}x_1x_2 - h_{32}y_1x_2
    \\
    y_2 = h_{21}x_1 + h_{22}y_1 + h_{23} - h_{31}x_1y_2 - h_{32}y_1y_2
\end{align*}

or more concisely:

\begin{equation*}
    \begin{bmatrix} x_1 & y_1 & 1 & 0 & 0 & 0 & -x_1x_2 & -y_1x_2 \\ 0 & 0 & 0 & x_1 & y_1 & 1 & -x_1y_2 & -y_1y_2   \end{bmatrix} \mathbf{h} = 0
\end{equation*}

where $\mathbf{h} = [h_{11},h_{12},h_{13},h_{21},h_{22},h_{23},h_{31},h_{32}]^{\top}$. We can stack such constraints for $n$ pair of points, leading to a system of equations of the form $\mathbf{A}\mathbf{h} = 0$ where $\mathbf{A}$ is a $2n \times 8$ matrix. Given the $8$ degrees of freedom and the system above, we need at least $8$ points (4 in each plan) to compute an estimation of our homography. This is the method we used to build the \domain{Homography Dataset}, and is also the method we use to compute the homography after our keypoints predictions. An example of the points correspondences in Soccer is available Figure \ref{supp:fig:eg-homography} \textbf{(right)}.

In reality, our Direct Estimation model doesn't predict directly the $8$ homography parameters, but the coordinates of $4$ control points instead \cite{jiang2019}. The predicted coordinates of these $4$ control points are then used to estimate the homography using the system of equations above.

\section{Supplementary datasets details}
\label{supp:sec:datasets}

\subsection{Overview}
\begin{center}
\begin{tabular}[p]{|l||c|c|c|c|c|c|c|}
    \hline
    \textbf{Name} &  
    \textbf{Inputs} & 
    \textbf{Outputs} & 
    \multiline{13mm}{\centering\textbf{\# Train \\ Images}} &
    \multiline{19mm}{\centering\textbf{\# Validation \\ Images}} &
    \multiline{13mm}{\centering\textbf{\# Test \\ Images}} &
    \textbf{Examples} \\\hline\hline
    \domain{Tracking Dataset} & $1024,1024$ & Bounding Boxes & 480 & 60 & 60 & Figure \ref{fig:eg-dataset} (a) \\\hline
    \domain{Homography Dataset} & $1280,720$ & $3,3$ & 874 & - & 159 & - \\\hline
    \domain{Keypoints Dataset} & $320,320$ & Keypoints &  463 & 46 & 59 &  Figure \ref{fig:eg-dataset} (b) \\\hline
\end{tabular}
\end{center}

\begin{figure*}[h]
  \centering
  \includegraphics[width=\textwidth]{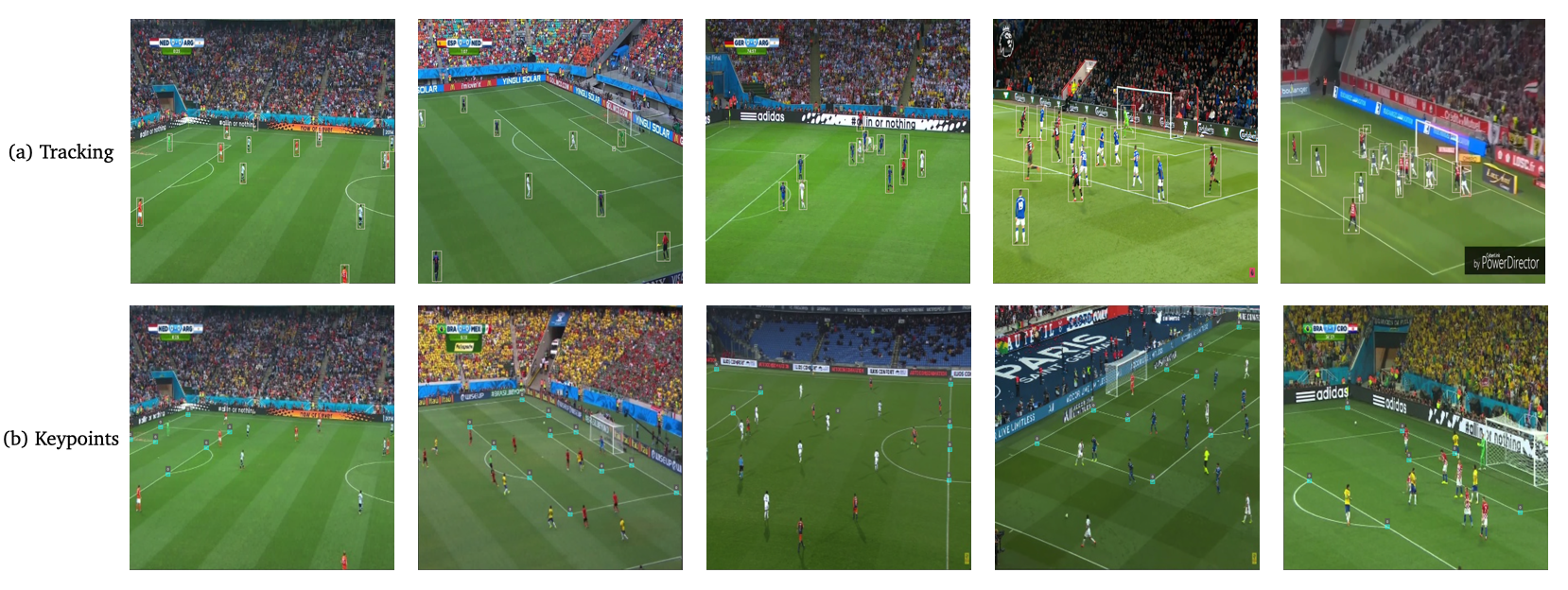}
  \caption{Example of the Tracking Dataset (a), with the bounding boxes for Players, Referees and the Ball. In (b) are shown examples of the Keypoints Dataset, with the position of the keypoints highlighted in blue.}
  \label{fig:eg-dataset}
\end{figure*}

\subsection{Actions for the DRL agent}

The controlled player can move in 8 directions, sprinting or not. These actions are sticky, and will last until an action \emph{Stop Moving} (resp. \emph{Stop Sprinting}) is produced. They can also produce several sort of passes or shots, and interact with dribbles or slides. The full list of actions is available Table \ref{fig:list-of-actions}.

\begin{table}[t!]
\begin{center}
\begin{tabular}[p]{lccc}

\toprule

Top & Bottom & Left & Right \\

Top Left & Top Right & Bottom Left & Bottom Right \\

Short Pass & High Pass & Long Pass & Shot \\

Do Nothing & Slide & Dribble & Stop Dribble \\

Sprint & Stop Moving & Stop Sprinting & - \\

\bottomrule
\end{tabular}
\caption{\label{fig:list-of-actions}List of actions the agent can chose at each timestep.}
\end{center}
\end{table}

\section{Supplementary results}
\label{supp:sec:results}

\subsection{Architectural choices for tracking}

In the following paragraph, we discuss some failed model experiments and architectural choices. Except for the entire model we presented above, we tested $4$ other different configurations: 
\newline 
\begin{enumerate*}
    \item Use a $2$-steps model for direct homography estimation,
    \item Use a Multi-scale network for direct homography estimation,
    \item Apply the tracking model to split images,
    \item Smooth the homographies estimations over time,
\end{enumerate*}
\;

For (1), we tried different scenarios based on a first direct homography estimation model. Giving the first model's prediction to a second one, also based on Resnet18, does not improve the performance significantly ($0.8$\%) (confirming previous results \cite{jiang2019}). We also tried to stack the field warped with our homography estimation to our original image, but without any improvements at all. Finally, using both the warped field and the homography estimation doesn't improve the performance either.

For (2), we implemented and trained a $3$-steps Multi-scale model from \cite{le2020deep}. This model progressively estimates and refines the homography from a third of the image to the entire one, in addition to the warped field from the previous step estimation. While promising, this model did not lead to better results. We believe that this is a result of the high degree of transformations there is between the field and the camera input. However, it can be a great model to estimate the homography modifications happening between $2$ consecutive frames. 

\begin{figure*}[h!]
  \centering
  \includegraphics[width=\textwidth]{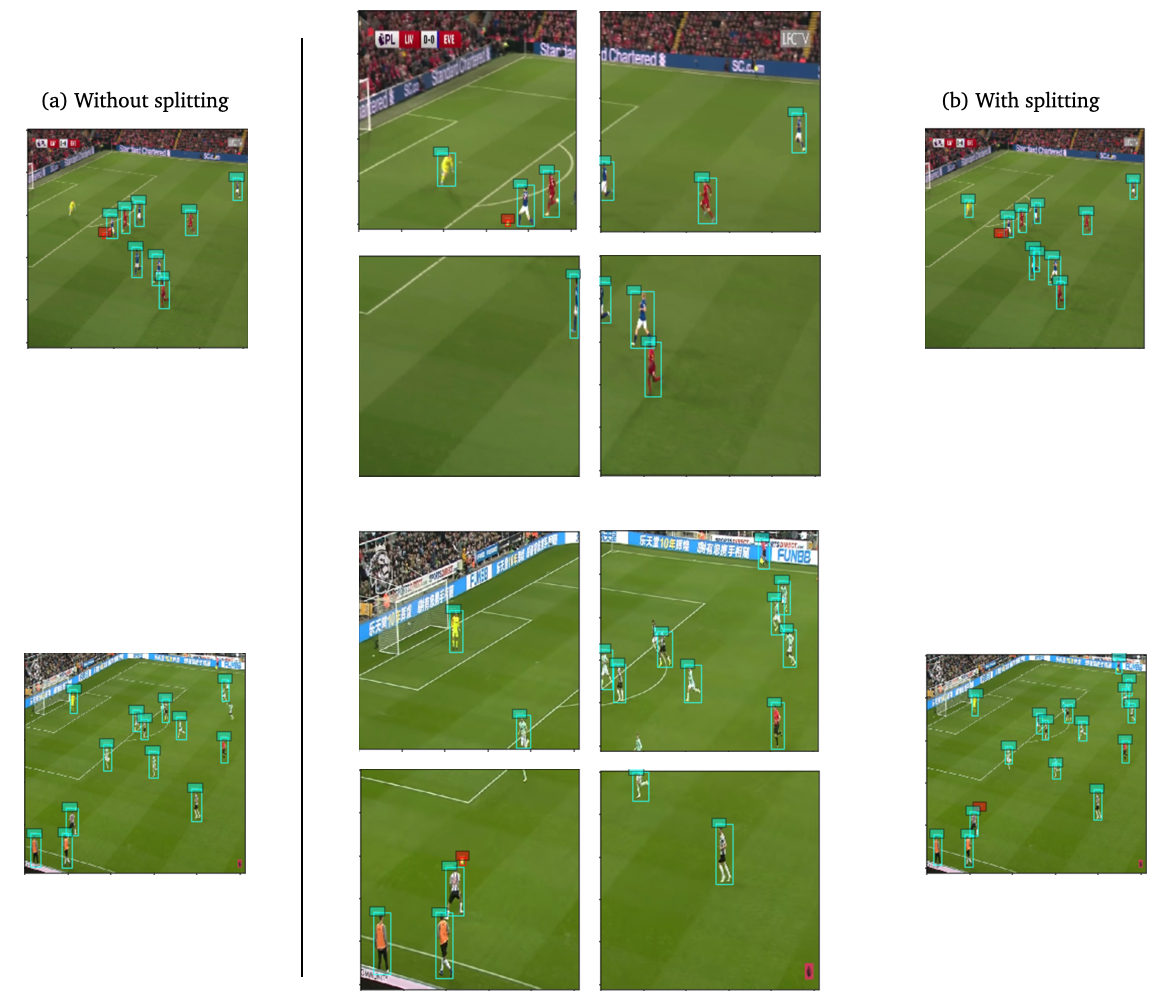}
  \caption{Example of the difference in results between a direct prediction, and a splitted prediction. It gives better on results on average, while letting some players (the cropped ones) being missed. This issue can be addressed by using both the direct prediction and the splitted one.}
  \label{fig:full-split}
\end{figure*}

For (3), we realized that producing the image in higher quality and then splitting it in separate panels allowed the tracking model to detect more players. An example is available Figure \ref{fig:full-split}. We compared the results of the tracking model on $3$ experiments: detection on the entire image (with shape ($1024,1024$), 4 detections on 4 panels (each of shape ($512,512$) and 8 detections on panels of shape ($256,256$). Overall, we found that the second experiment was leading to the best results.

For (4), we tried to apply a Sagvol filter to the sequence of estimated homographies over the entire action. However, the results were mitigated, as it would help the stability over time, but would completely ruin the tracking as soon as one outlier was produced. To avoid such behaviour, we ended up using no filter to the sequence of homographies.

\begin{figure}[h!]
  \centering
  \includegraphics[width=.55\textwidth]{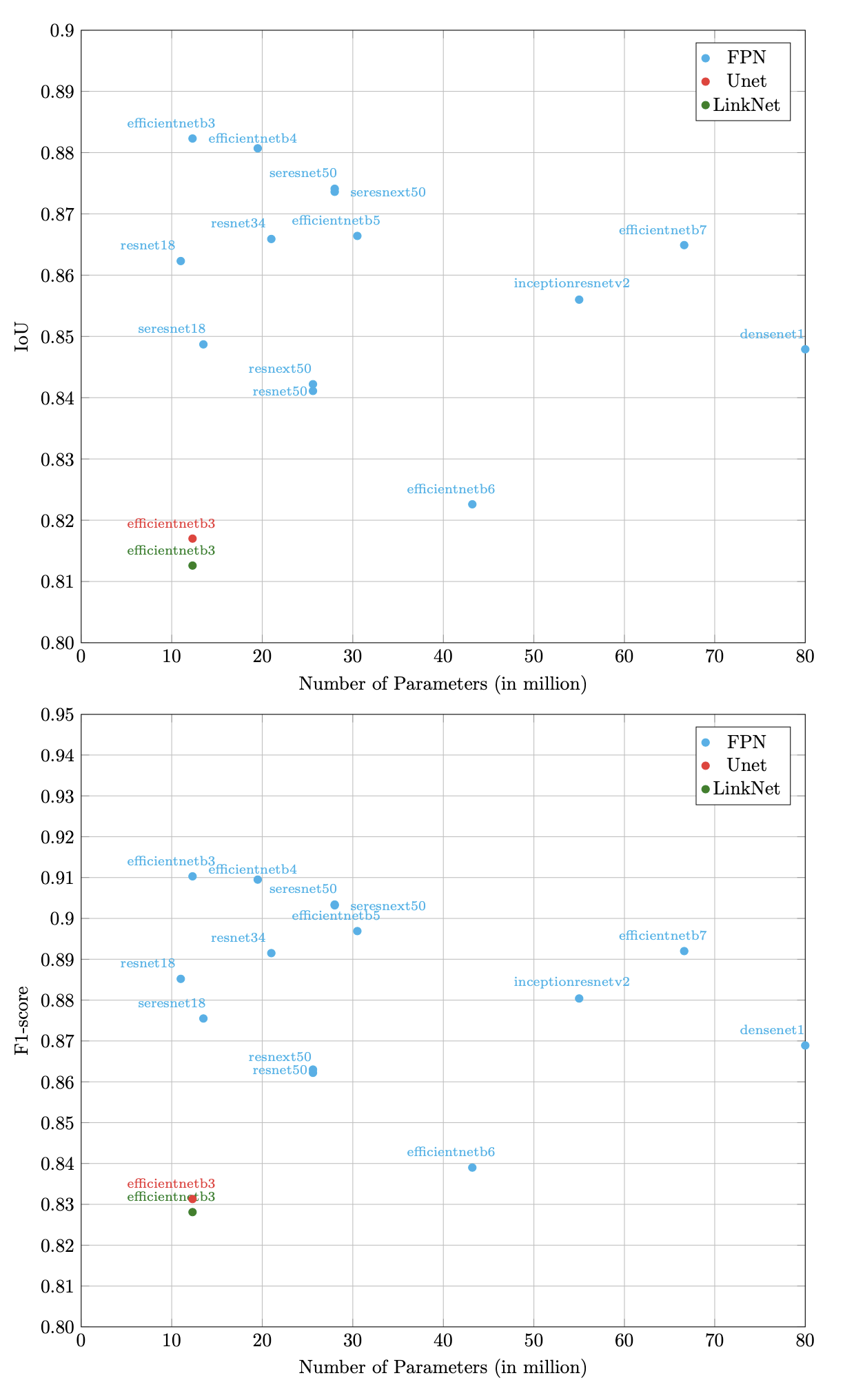}
  \caption{Ablation study on the keypoints dataset to generate masks. Each model is plotted based on its number of parameters and its metric. We only tested one backbone for the LinkNet et Unet architecture because of the poor results in comparison to the FPN.}
  \label{supp:fig:keypoints-ablation}
\end{figure}

\subsection{Expected Discounted Goal: hyperparameters}
\label{supp:sec:drl-hyp}
\begin{table}[h!]
\begin{center}
\begin{tabular}[p]{lc}

\toprule

\textbf{Parameter} & \textbf{Value} \\

\hline 

Clipping Range & 0.08 \\
Discount Factor &  0.993 \\
Entropy Coefficient & 0.003 \\
GAE & 0.95 \\
Gradient Norm Clipping & 0.64 \\
Learning Rate & .000343 \\
Number of Actors & 16\\
Training Epochs per Update & 2 \\
Value Function Coefficient & 0.5 \\

\bottomrule
\end{tabular}
\caption{Hyperparameters used for each global step of the Agent training. We reused these for selfplay too.}
\end{center}
\end{table}

\subsection{Tracking and homography: supplementary results}
\label{supp:sec:drl-supp-results}

\begin{figure*}[h]
  \centering
  \includegraphics[width=\textwidth]{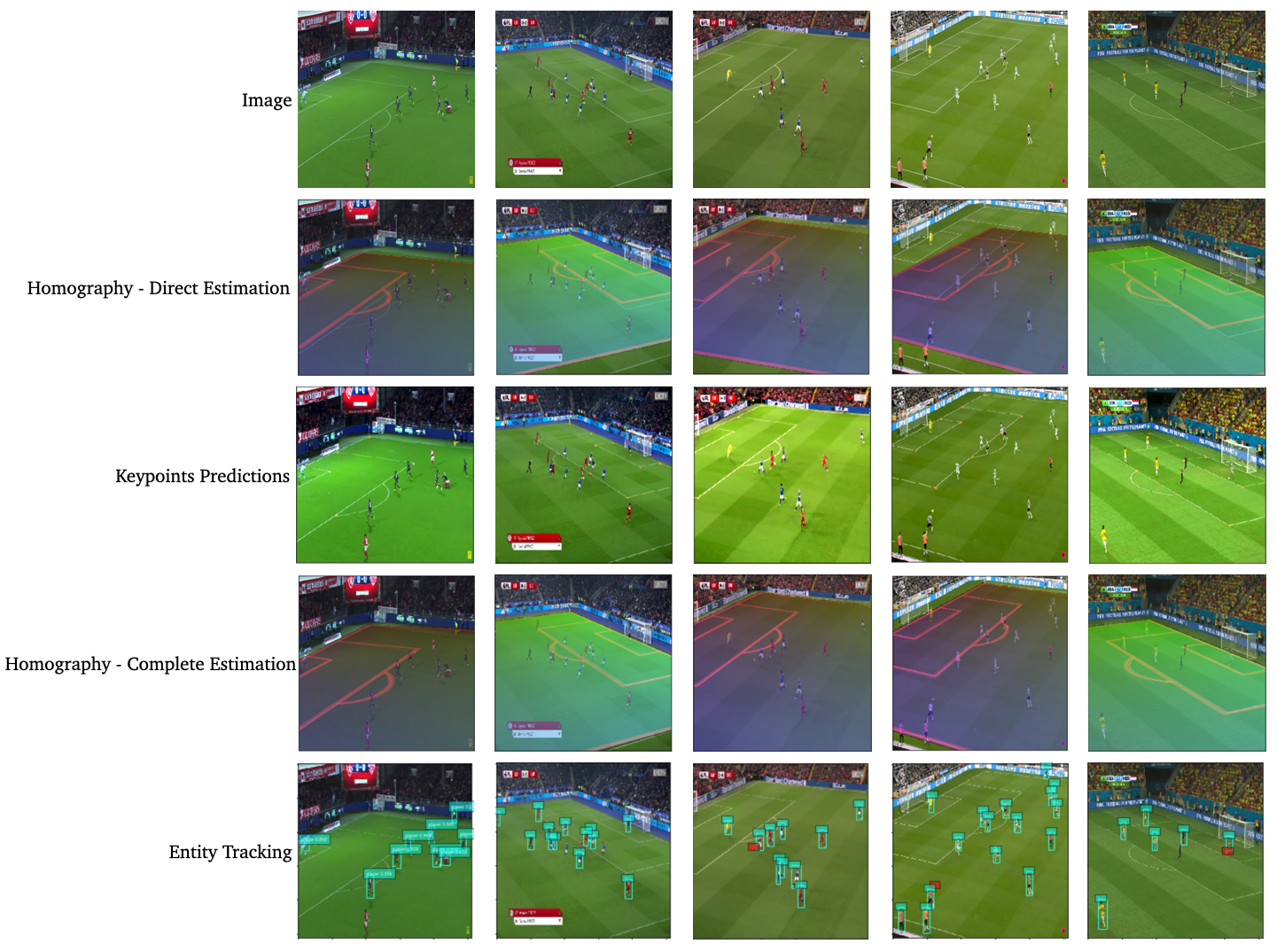}
  \caption{Additional results from each model of our tracking model. Each image was selected randomly and not cherry picked.}
  \label{fig:full-results}
\end{figure*}

See Figure \ref{fig:full-results} for an overview of the results for each model on random images.

\subsection{Expected Discounted Goal: supplementary results}

\begin{figure}[h!]
  \centering
  \includegraphics[width=.55\textwidth]{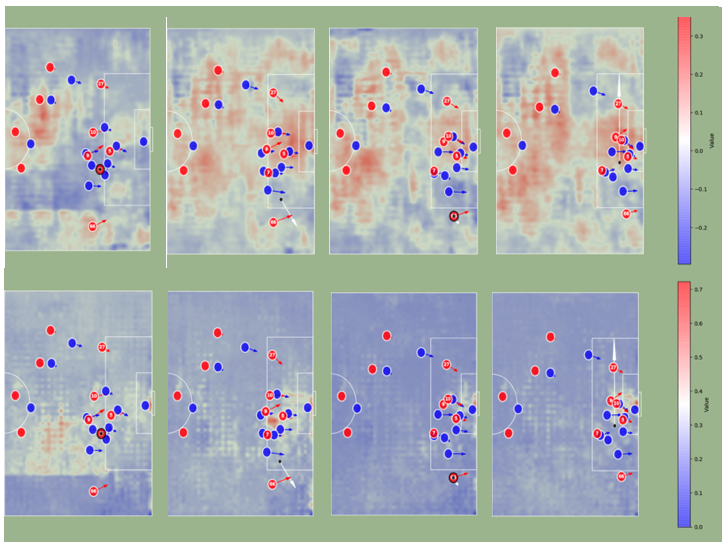}
  \caption{Another example following the idea from Figure \ref{fig:res-value} (right). Again, we show here how the easy Agent doesn't grasp any meaningful information from the game.}
  \label{supp:fig:us-vs-easy}
\end{figure}

See Figure \ref{supp:fig:us-vs-easy} for another example of our agent versus an agent trained only against an easy bot.

\end{document}